\documentclass[11pt]{article}

\usepackage[letterpaper,left=1.15in,right=1.05in,top=0.75in,bottom=0.8in]{geometry}
\usepackage{fontspec}
\newfontfamily\arabicfont[Script=Arabic,
  BoldFont=Amiri-Bold.ttf,
  ItalicFont=Amiri-Italic.ttf,
  BoldItalicFont=Amiri-BoldItalic.ttf
]{Amiri-Regular.ttf}
\newcommand{\textarabic}[1]{%
  \begingroup\arabicfont\beginR #1\endR\endgroup}
\usepackage{microtype}

\usepackage{amsmath,amssymb,amsthm}
\usepackage{graphicx}
\usepackage{booktabs}
\usepackage{tabularx}
\usepackage{colortbl}
\usepackage{pdflscape}
\usepackage{placeins}
\usepackage{float}
\usepackage{natbib}
\usepackage{xcolor}
\definecolor{paperlinkblue}{RGB}{0,0,200}

\usepackage[
  colorlinks=true,
  linkcolor=paperlinkblue,
  citecolor=paperlinkblue,
  urlcolor=paperlinkblue
]{hyperref}
\usepackage[nameinlink,noabbrev]{cleveref}
\usepackage{fancyhdr}

\setcitestyle{authoryear,square,comma,aysep={,}}
\fancypagestyle{paper}{%
  \fancyhf{}
  \fancyfoot[C]{\small\thepage}
  
  }
\fancypagestyle{firstpage}{%
  \fancyhf{}
  
  }
\graphicspath{{figures/}}

\theoremstyle{definition}

\theoremstyle{remark}

\makeatletter
\renewcommand{\maketitle}{%
  \thispagestyle{firstpage}%
  \vspace*{-1.2em}%
  \noindent\rule{\textwidth}{5pt}\par
  \vspace{1.25em}%
  {\raggedright\fontsize{19}{23}\selectfont\bfseries\@title\par}%
  \vspace{1.1em}%
  \noindent\rule{\textwidth}{0.8pt}\par
  \vspace{2.7em}%
  \begin{center}
    {\large\bfseries\@author\par}
  \end{center}
  \vspace{2.0em}%
}
\makeatother

\renewenvironment{abstract}{%
  \begin{center}
    {\large\bfseries Abstract\par}
    \vspace{0.8em}
    \begin{minipage}{0.84\textwidth}
      \small
}{%
    \end{minipage}
  \end{center}
  \vspace{0.9em}
}

\title{%
  \textbf{HalluTruthQA-4K: A Fine-Grained Corpus }\\[1mm]
  \textbf{ and Annotation Process for Arabic}\\
  \textbf{ Hallucination Detection and Truth Verification
}%
}

\author{%
  Salah Eddine Bekhouche$^{ 1}$ \quad
  Abdessalam Bouchekif$^{ 2}$ \quad
  Hichem Telli$^{ 3}$ \\
  Mohammed-En-Nadhir Zighem \quad
  Abdenour Hadid$^{ 4}$\\[2mm]
  \small $^{ 1}$University of the Basque Country, Spain \quad
  $^{2}$Hamad Bin Khalifa University, Qatar \\
  \small $^{3}$University of Biskra, Algeria \quad
  $^{4}$Universiti Malaysia Kelantan, Malaysia
}

\date{}

\begin{document}

\maketitle

\begin{abstract}
Large language models can generate fluent answers, yet factual errors remain challenging to detect, localize, explain, and verify. Existing Arabic hallucination benchmarks typically focus on response-level judgments, with limited support for identifying the exact erroneous content, explaining why it is incorrect, or identifying the correct factual answer. We introduce \textsc{HalluTruthQA-4K}, an expert-annotated extension of the \textsc{HalluTruthQA} benchmark, expanding the corpus from 2,400 to 4,000 Arabic question answering (QA) instances across four knowledge-intensive domains: Islamic knowledge, history, science, and geography. Each instance contains an Arabic question, a model-generated response, a verified reference answer, and six candidate answers consisting of one correct answer and five plausible distractors. For hallucinated responses, the corpus additionally provides character-level hallucination spans and human-written explanations. The resulting corpus contains 1,789 hallucinated and 2,211 non-hallucinated responses, with 2,419 annotated erroneous spans. These annotations support four main evaluation tasks: binary hallucination detection, hallucination span localization, explanation generation and evaluation, and multiple-choice factual verification (MCQ). We describe the corpus construction and quality-control process, including controlled response generation, expert annotation, independent verification, and adjudication. By providing multiple levels of factual error annotation within the same QA instances, \textsc{HalluTruthQA-4K} enables fine-grained evaluation of factual reliability in Arabic LLMs. The corpus is publicly available on Hugging Face\footnote{\url{https://huggingface.co/datasets/Bekhouche/HalluTruthQA-4K}}.

\end{abstract}
\begin{center}
\end{center}
\vspace{0.5em}

\section{Introduction}
\label{sec:introduction}

Large language models (LLMs) can generate fluent and convincing responses,
but fluency does not guarantee factual correctness. Hallucination generally
refers to generated content that is factually incorrect, fabricated,
unsupported, misleading, or inconsistent with the available context
\citep{ji2023survey,huang-etal-2025-survey}. In knowledge-intensive
question answering (QA), hallucinations may involve incorrect entities,
dates, numerical values, quotations, source attributions, or explanations.
A response may therefore appear coherent and confident while still
containing factual errors that reduce its reliability.

Hallucinations can arise from incomplete or noisy training data, limited
coverage of domain-specific knowledge, ambiguous prompts, or insufficient
contextual grounding. They are also related to the next-token prediction
objective of LLMs, which favors probable continuations without explicitly
verifying the factual correctness of each generated claim
\citep{huang-etal-2025-survey,alansari2026survey}. This challenge is
particularly important for Arabic, where high-quality, diverse, and
domain-specific evaluation resources remain limited.

Recent Arabic benchmarks have evaluated factual knowledge in Islamic
studies, legal reasoning, and general culture
\citep{qias2025,alwajih2025palmx,abdelaal2026islamicmmlu,qias2026,
bouchekif2026mawarith,gaben2026islamicturathbench}. Many of these
benchmarks use a multiple-choice format, which is useful for measuring
factual knowledge but provides limited information about errors in
free-form generation, such as fabricated claims, incorrect quotations,
or misleading explanations. Other Arabic resources study hallucination
and factual reliability in sentence-level evaluation, generative QA,
summarization, and domain-specific quotation verification
\citep{mubarak2024halwasa,alansari2025arahallueval,
alansari2026halluscore,mubarak2025islamiceval}. However, response-level
labels, error spans, explanations, and verified answers are often provided
in separate datasets rather than jointly within the same QA instances.

A response-level hallucination label is useful for detection, but it does
not show where the error occurs or why it is incorrect. Two responses with
the same hallucination label may contain very different types of errors.
One response may be entirely incorrect, while another may provide the
correct main answer but include an incorrect date, a fabricated quotation,
or a false supporting statement. Similarly, a model may correctly identify
a response as hallucinated without locating the erroneous text, explaining
the error, or selecting the correct factual answer. A fine-grained
evaluation framework should therefore address four questions:
\textit{Does the response contain a hallucination? Where does the error
occur? Why is it incorrect? What is the verified answer?}

These distinctions are especially important in Arabic knowledge-intensive
QA because verification requirements differ across domains. Islamic
knowledge requires careful verification of Qur'anic verses, hadith,
scholarly opinions, and source attributions. Historical questions require
accurate chronology and correct identification of people, places, and
events. Scientific questions often depend on precise definitions,
numerical values, measurements, and causal relations. Geography requires
accurate entities, locations, borders, and spatial relations. Few existing
Arabic resources jointly provide response-level hallucination detection,
exact character-level error localization, human-written explanations,
fine-grained error types, and factual answer verification within the same
QA instances.

We previously introduced \textsc{HalluTruthQA}, a benchmark containing
2,400 expert-annotated Arabic QA instances across Islamic knowledge,
history, science, and geography
\citep{bouchekif2026hallutruthqa}. The benchmark combines response-level
hallucination labels, character-level erroneous spans, human-written
explanations, verified reference answers, a hierarchical hallucination
taxonomy, and six candidate answers for factual verification. These
annotation layers support different evaluation capabilities, including
hallucination detection, error localization, explanation, and factual
answer selection.

In this work, we introduce \textsc{HalluTruthQA-4K}, extending the
original benchmark from 2,400 to 4,000 expert-annotated Arabic QA
instances, with 1,000 examples in each domain. All free-form responses
were generated using \textsc{Fanar-1-9B-Instruct}
\citep{fanarteam2025fanar}, an Arabic-centric instruction-tuned LLM,
under a controlled generation setting. The extended corpus preserves the
fine-grained annotation framework of \textsc{HalluTruthQA} while
increasing its scale and providing balanced training, development, and
test splits.

The main contributions of this work are as follows:
\vspace{-\parskip}
\begin{itemize}
    \setlength{\topsep}{0pt}
    \setlength{\partopsep}{0pt}
    \setlength{\itemsep}{0pt}
    \setlength{\parsep}{0pt}

    \item We extend \textsc{HalluTruthQA} from 2,400 to 4,000
    expert-annotated Arabic QA instances, with 1,000 examples in each
    of four knowledge-intensive domains: Islamic knowledge, history,
    science, and geography. The corpus supports binary hallucination
    detection, hallucination span localization, explanation generation
    and evaluation, and multiple-choice factual verification.

    \item We describe the corpus construction and quality-control
    process, including question preparation, controlled response
    generation, candidate-answer construction, expert annotation,
    independent verification, adjudication, the hallucination taxonomy,
    and the released data format.
\end{itemize}
\vspace{-\parskip}

\noindent
\textsc{HalluTruthQA-4K} also serves as the official dataset for
Track~2 of the HalluScoring 2026 shared task. Track~2.1 focuses on
Islamic knowledge, while Track~2.2 covers history, science, and geography.

\section{Related Work}
\label{sec:related_work}

Hallucination evaluation has been studied across question answering,
summarization, retrieval-augmented generation, and long-form generation.
Existing benchmarks differ in the types of errors they target, the
information available for verification, and the granularity of their
annotations \citep{huang2025survey}. Several widely used benchmarks
illustrate these different evaluation settings. \textsc{TruthfulQA}
\citep{lin2022truthfulqa} uses adversarial questions designed to elicit
common misconceptions, while \textsc{HaluEval}
\citep{li2023halueval} combines automatically generated and human-annotated
examples for hallucination evaluation. \textsc{FELM}
\citep{chen2023felm} provides fine-grained factual error annotations across
multiple domains, whereas \textsc{RAGTruth} \citep{niu2024ragtruth}
focuses on hallucinations in retrieval-augmented generation.
\textsc{FActScore} \citep{min2023factscore} decomposes long-form
generations into atomic factual claims, and \textsc{FAVA}
\citep{mishra2024fava} addresses fine-grained hallucination detection and
correction at the span level. Together, these benchmarks highlight the
importance of moving beyond a single response-level judgment toward more
localized analysis of factual errors.

Multilingual benchmarks have also explored span- and type-level
hallucination annotations. Mu-SHROOM \citep{vazquez2025mushroom}
formulates multilingual hallucination detection as a span-labeling task
and includes Modern Standard Arabic. Follow-up work has explored
semantic-role decomposition and textual entailment for this benchmark
\citep{elchafei2025hallucination}. \textsc{HalluVerse-M$^3$}
\citep{abdaljalil2026halluverse} covers Arabic, English, Hindi, and Turkish
and supports controlled comparisons across question answering and dialogue
summarization. These benchmarks provide useful settings for multilingual
hallucination evaluation, but they are not specifically designed for
knowledge-intensive Arabic question answering.


\begin{landscape}
\begin{table}[!ht]
\centering
\caption{Comparison of \textsc{HalluTruthQA-4K} with recent Arabic and
multilingual hallucination resources. \emph{Partial} indicates that a
feature is available only for a subset of the data.}
\label{tab:hallucination_resources}

\footnotesize
\setlength{\tabcolsep}{2.3pt}
\renewcommand{\arraystretch}{1.22}

\begin{tabularx}{0.96\linewidth}{@{}
>{\raggedright\arraybackslash}p{2.80cm}
>{\raggedright\arraybackslash}p{1.70cm}
>{\raggedright\arraybackslash}p{2.10cm}
>{\raggedright\arraybackslash}p{1.70cm}
>{\raggedright\arraybackslash}X
>{\raggedright\arraybackslash}X
>{\raggedright\arraybackslash}X
>{\raggedright\arraybackslash}X
>{\raggedright\arraybackslash}X@{}}

\toprule
\rowcolor{black!8}
\textbf{Resource} &
\textbf{Languages} &
\textbf{Domain / task} &
\textbf{Outputs} &
\textbf{Human review} &
\textbf{Span information} &
\textbf{Types or labels} &
\textbf{Explanations} &
\textbf{Verified answer / correction} \\
\midrule

\textsc{AraHalluEval}\newline
\citeyearpar{alansari2025arahallueval}
& Arabic
& Generative QA and abstractive summarization
& LLM outputs
& Two annotators and expert adjudication
& No exact error spans
& Twelve factuality and faithfulness indicators
& No error-specific explanations
& Reference answer for QA or source document for summarization \\
\addlinespace[3pt]

\textsc{HalluVerse-M$^3$}\newline
\citeyearpar{abdaljalil2026halluverse}
& Arabic, English, Hindi, and Turkish
& QA and dialogue summarization
& Controlled LLM-edited outputs
& Human validation
& No exact character offsets
& Entity-, relation-, and sentence-level hallucination types
& No
& Original ground-truth output \\
\addlinespace[3pt]

\textsc{HalluScore}\newline
\citeyearpar{alansari2026halluscore}
& Arabic
& Multi-domain generative QA
& LLM outputs
& Human annotation of model responses
& Partial localization for partially hallucinated responses
& Factual, faithfulness, and partial-hallucination labels
& Generated and subsequently reviewed
& Verified answer and supporting source \\
\addlinespace[3pt]

\textsc{Aftina}\newline
\citeyearpar{mohammed2025aftina}
& Arabic
& Islamic fatwa QA
& RAG-supported LLM outputs
& Expert evaluation of generated outputs
& No
& No corpus-level hallucination taxonomy
& No
& Dar Al-Ifta answers and retrieved supporting content \\
\addlinespace[3pt]

\textsc{IslamicEval}\newline
\citeyearpar{mubarak2025islamiceval}
& Arabic
& Qur'an and Hadith
& LLM outputs
& Islamic-studies experts
& Complete quoted-passage spans
& Correct or incorrect quotation labels
& No
& Canonical correction or indication that no valid correction exists \\
\addlinespace[3pt]

\textsc{HalluTruthQA}\newline
\citeyearpar{bouchekif2026hallutruthqa}
& Arabic
& Islamic knowledge, history, science, and geography
& LLM outputs
& Domain experts and independent verification
& Exact character-level erroneous spans
& Macro- and micro-level hallucination types
& Human-written explanations of localized errors
& Verified reference answer and six candidate answers \\
\midrule

\rowcolor{blue!5}
\textbf{\textsc{HalluTruthQA-4K} (ours)}
& \textbf{Arabic}
& \textbf{Islamic knowledge, history, science, and geography}
& \textbf{LLM outputs}
& \textbf{Domain experts and independent verification}
& \textbf{Exact character-level erroneous spans}
& \textbf{Macro- and micro-level hallucination types}
& \textbf{Human-written explanations of localized errors}
& \textbf{Verified reference answer and six candidate answers} \\

\bottomrule
\end{tabularx}

\end{table}
\end{landscape}
Arabic-specific hallucination evaluation has recently received increasing
attention. \textsc{Halwasa} \citep{mubarak2024halwasa} studies
hallucinations in Arabic LLM-generated text using isolated sentences
conditioned on predefined keywords rather than natural question--answer
interactions. \textsc{AraHalluEval}
\citep{alansari2025arahallueval} introduces a multidimensional manual
evaluation framework for Arabic generative QA and abstractive summarization.
It uses 12 factuality and faithfulness indicators covering errors involving
entities and numerical values, contradictions, source conflicts, and
fabrications. \textsc{HalluScore}
\citep{alansari2026halluscore} focuses on hallucination-prone Arabic
questions across multiple domains, cultural settings, and reasoning
requirements. It provides verified evidence, human response-level
annotations, fine-grained labels, and reviewed answer explanations. These
resources support analysis beyond binary hallucination detection. However,
neither provides systematic character-level localization of every erroneous
segment in hallucinated QA responses. In \textsc{HalluScore}, localized
hallucinated text is provided for partially hallucinated responses rather
than as character-offset annotations for every erroneous segment.

Domain-specific Arabic resources have also examined factual reliability in
Islamic content. \textsc{Aftina} \citep{mohammed2025aftina} investigates
hallucination mitigation in fatwa generation using retrieval-augmented
generation and reranking. \textsc{IslamicEval}
\citep{mubarak2025islamiceval} evaluates the identification, validation,
and correction of Qur'anic and Hadith quotations. Although it includes span
annotations, these spans identify complete intended quotations rather than
the exact erroneous characters or segments within the generated response.
More recently, \textsc{IslamicFaithQA}
\citep{bhatia2026islamicfaithqa} introduced a 3,810-item bilingual
Arabic--English benchmark with atomic single-gold answers for evaluating
hallucination and abstention. These resources provide important evaluation
settings for source-sensitive Islamic QA, but their domain-specific scope
differs from the multi-domain setting considered here.

Table~\ref{tab:hallucination_resources} compares closely related Arabic
and multilingual hallucination resources.

\section{Data Description}
\label{sec:data}

\textsc{HalluTruthQA-4K} contains 4{,}000 Arabic question answering (QA)
instances covering four knowledge-intensive domains: Islamic knowledge,
history, science, and geography. The corpus is balanced across domains,
with 1{,}000 instances per domain. It extends the original 2{,}400-instance
\textsc{HalluTruthQA} benchmark while preserving the same domains and
annotation framework. The final release contains 1{,}600 training,
800 development, and 1{,}600 test instances.

Each instance contains an Arabic question, a model-generated free-form
response, an expert-verified reference answer, a binary hallucination label,
and six candidate answers for factual verification. For hallucinated
responses, the dataset additionally provides one or more character-level
erroneous spans, each associated with a human-written explanation.

The four domains were selected to represent different factual verification
challenges. Islamic knowledge requires careful verification of source
attributions and supporting evidence. History emphasizes entities, events,
and chronology. Science requires accurate concepts, numerical values,
measurements, and causal relations. Geography focuses on entities,
locations, quantitative facts, and spatial relations. 
The dataset is split into 400 training, 200 development, and 400 test
instances per domain.
Table~\ref{tab:label_span_distribution} summarizes the distribution of
hallucination labels and localized erroneous spans across domains.  We also checked split disjointness using the
\texttt{(id, question)} pair and found no duplicated pairs across the
released splits. The 1{,}600-instance test set is also used as the official
Track~2 test set of the HalluScoring 2026 shared task.

\begin{figure}[!h]
    \centering
    \includegraphics[width=0.92\linewidth]{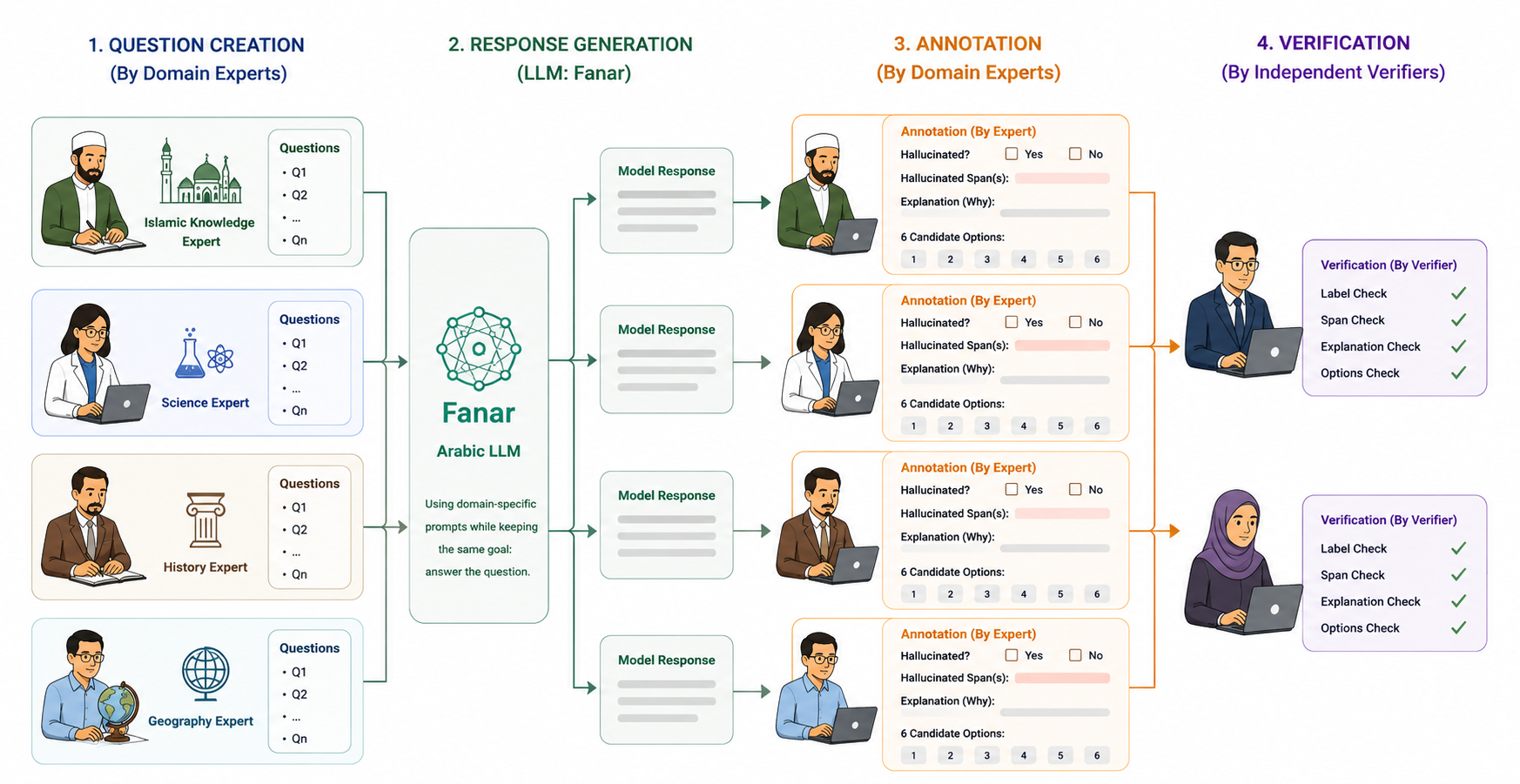}
    \caption{Construction and annotation pipeline of
    \textsc{HalluTruthQA-4K}. Domain experts prepare the questions and
    verify the reference answers. \textsc{Fanar-1-9B-Instruct} generates
    the free-form responses, after which candidate answers are constructed.
    The resulting instances are annotated for hallucination labels,
    erroneous spans, and explanations, followed by independent verification
    and expert adjudication.}
    \label{fig:hallutruthqa_construction}
\end{figure}

\begin{table}[t]
\centering
\small
\setlength{\tabcolsep}{6pt}
\renewcommand{\arraystretch}{1.12}
\caption{Distribution of hallucination labels and localized erroneous spans
across the four domains.}
\label{tab:label_span_distribution}

\begin{tabular}{lcccc}
\toprule
\textbf{Domain} &
\textbf{Hallucinated} &
\textbf{Non-hallucinated} &
\textbf{Erroneous spans} &
\textbf{Total} \\
\midrule

Islamic knowledge
& 497 (49.7\%)
& 503 (50.3\%)
& 686
& 1{,}000 \\

History
& 353 (35.3\%)
& 647 (64.7\%)
& 399
& 1{,}000 \\

Science
& 464 (46.4\%)
& 536 (53.6\%)
& 528
& 1{,}000 \\

Geography
& 475 (47.5\%)
& 525 (52.5\%)
& 806
& 1{,}000 \\

\midrule
\rowcolor{gray!12}
\textbf{Total}
& \textbf{1{,}789 (44.7\%)}
& \textbf{2{,}211 (55.3\%)}
& \textbf{2{,}419}
& \textbf{4{,}000} \\

\bottomrule
\end{tabular}
\end{table}
\subsection{Corpus Construction and Annotation}
\label{sec:annotation}
The construction of \textsc{HalluTruthQA-4K} follows a four-stage pipeline:
question preparation, response generation, expert annotation, and independent
verification. Figure~\ref{fig:hallutruthqa_construction} summarizes the
complete process. Domain experts first prepare and verify the questions and
reference answers. The responses are then generated using
\textsc{Fanar-1-9B-Instruct}, after which the instances are annotated for
hallucination labels, erroneous spans, explanations, and candidate answers.
Finally, an independent verification pass is performed before final
adjudication.
\paragraph{Question preparation and reference verification.}

Questions were manually prepared by domain experts using their subject
knowledge and relevant reference materials. Experts were instructed to
write clear, knowledge-intensive questions while maintaining diversity in
topic and difficulty. Each reference answer was verified against reliable
sources from the corresponding domain. Questions that were ambiguous,
underspecified, or compatible with multiple defensible answers were revised
or removed.  Verification relied on a range of domain-specific sources. For Islamic
knowledge, these included established online resources and Arabic reference
works covering Qur'anic studies, creed, Prophetic biography, and related
topics. Questions in history, science, and geography were checked against
relevant books, atlases, encyclopedias, and other reference materials. A
representative list of the consulted sources is provided in
Appendix~\ref{app:sources}.

Reference verification was based on factual meaning rather than exact
string matching. Semantically equivalent answers were accepted despite
differences in wording, transliteration, or formatting. Equivalent Hijri
and Gregorian dates were also accepted when they referred to the same
historical event.

\paragraph{Response generation.}

All responses were generated using
\textsc{Fanar-1-9B-Instruct}\footnote{
\url{https://huggingface.co/QCRI/Fanar-1-9B-Instruct}}
\citep{fanarteam2025fanar}. We use a single Arabic-centric generator to
maintain a controlled generation setting and avoid additional variation
caused by differences in model architecture, capability, and response
style. The model was used only to generate the responses. Reference
verification, hallucination annotation, span localization, explanation
writing, candidate-answer construction, and answer-key validation were
performed manually. Generation used a maximum of 1{,}024 new tokens,
temperature 0.0, top-$p$ of 1.0, batch size 4, and \texttt{bfloat16}
precision on a 48\,GB GPU.

\paragraph{Hallucination labeling.}

A response was labeled as \texttt{hallucination} if it contained at least
one factually incorrect, fabricated, unsupported, misleading, or
contextually inconsistent claim. Annotation considered the complete response rather than focusing only on
directly answering the question. Indeed, a response was labeled
as hallucinated even when its main answer was correct if it also contained
an incorrect quotation, date, numerical value, attribution, citation, or
supporting explanation.

A response was labeled as \texttt{no\_hallucination} when its factual
claims were consistent with the verified reference answer and the
consulted sources. Differences in wording, transliteration, level of
detail, or date format were accepted when they preserved the same factual
meaning.

\paragraph{Span annotation and explanations.}

For each hallucinated response, annotators identified one or more
character-level spans corresponding to the erroneous content. Span
selection followed a \emph{minimal but complete} principle: the selected
text had to contain enough information to identify the factual error while
excluding unrelated correct content.

When a response contained multiple independent errors, each error was
annotated separately. For example, an incorrect Qur'anic reference and an
independently fabricated quotation were represented as two distinct spans.
Each span was paired with a human-written explanation describing why the
selected content was incorrect or unsupported and providing the factual
correction when available. For cases in which the complete response was unusable, nonsensical, or
irrelevant to the question, the full response was annotated as the
erroneous span because no shorter segment adequately represented the
failure.

\begin{table}[H]
\centering
\small
\setlength{\tabcolsep}{5pt}
\renewcommand{\arraystretch}{1.12}
\caption{Annotation decisions for representative boundary cases.}
\label{tab:annotation_decisions}
\begin{tabularx}{\linewidth}{@{}
>{\raggedright\arraybackslash}p{0.38\linewidth}
>{\raggedright\arraybackslash}X@{}}
\toprule
\textbf{Case} & \textbf{Annotation decision} \\
\midrule

Main answer correct, but supporting evidence incorrect
& Label the response as hallucinated and annotate only the erroneous
evidence. \\

Semantically equivalent wording
& Do not label as hallucinated when the factual meaning is preserved. \\

Equivalent Hijri and Gregorian dates
& Do not label as hallucinated when both dates identify the same event. \\

Two independent errors
& Annotate two separate minimal-but-complete spans, each with its own
explanation and, where available, type. \\

Entire response unusable, nonsensical, or globally irrelevant
& Label the response as hallucinated and annotate the complete response
as the erroneous span. \\

\bottomrule
\end{tabularx}
\end{table}

\paragraph{Candidate-answer construction.}

Each question was paired with six manually constructed candidate answers:
one verified correct answer and five plausible distractors. We used six
options to reduce the probability of obtaining the correct answer by random
guessing to $1/6$ (16.7\%). Distractors were factually incorrect but
semantically related to the question and were designed to be similar to
the correct answer in style and level of specificity, limiting the use of
superficial cues for answer selection.  The candidate sets were constructed and validated by the domain experts and
were  independently reviewed by the trained research assistants.
They verified that each question had exactly one defensible correct option
and that no distractor constituted a semantically valid alternative to the
verified answer.

\paragraph{Verification and adjudication.}

The annotation process was conducted in two stages. In the first stage,
four domain experts, one for each domain, verified the questions and
reference answers, assigned binary hallucination labels, identified
erroneous spans, wrote explanations, and constructed and validated the
candidate answers and answer keys.
In the second stage, two trained research assistants, who were not involved
in the initial annotation, independently reviewed the annotated records.
Their verification covered the questions, reference answers, hallucination
labels, span boundaries, explanations, candidate answers, and answer keys.
. Cases involving disagreement, uncertainty, or inconsistent annotations were
returned to the corresponding domain expert for final adjudication.

Because the four experts annotated separate domain-specific subsets,
agreement was measured between the initial expert annotations and the
independent verification pass rather than among the four experts (see Table \ref{tab:annotation_reliability}).
Agreement was computed before final adjudication using Cohen's $\kappa$ for
categorical annotations, exact agreement for the multiple-choice answer key,
character-level F1 for span localization, and Intersection-over-Union (IoU)
for span overlap.

\begin{table}[H]
\centering
\small
\setlength{\tabcolsep}{5pt}
\renewcommand{\arraystretch}{1.10}
\caption{Expert--reviewer agreement computed before final adjudication.
Categorical type-agreement scores refer to records carrying taxonomy
annotations.}
\label{tab:annotation_reliability}
\begin{tabular}{@{}llc@{}}
\toprule
\textbf{Annotation dimension} &
\textbf{Metric} &
\textbf{Score} \\
\midrule

Binary hallucination label &
Cohen's $\kappa$ &
0.92 \\

Macro hallucination type &
Cohen's $\kappa$ &
0.86 \\

Micro hallucination type &
Cohen's $\kappa$ &
0.80 \\

Multiple-choice answer key &
Exact agreement &
0.976 \\

Span localization &
Character-level F1 &
0.88 \\

Span overlap &
IoU &
0.83 \\

\bottomrule
\end{tabular}
\end{table}

The agreement results indicate strong consistency across annotation
dimensions. Agreement is highest for the binary hallucination label,
indicating that experts and reviewers generally agreed on whether a response
contained a factual error. Agreement is lower for fine-grained type
classification, where distinctions between categories such as unsupported
claims, citation mismatches, wrong source attributions, and over-specific
additions require more detailed judgment.


\subsection{Data Format}
\label{sec:data_format}

Each instance is stored as a structured JSON record. Character offsets are
zero-based and \texttt{span\_end} is exclusive. For hallucinated responses,
the \texttt{hallucinations} field contains one or more localized-error
objects. This field is omitted for non-hallucinated responses. Table~
\ref{tab:data_fields} summarizes the fields in the final release. Domain
membership is encoded by the instance identifier prefix and by the split
construction rather than by a separate \texttt{domain} field.

\begin{table}[h]
\centering
\small
\setlength{\tabcolsep}{6pt}
\renewcommand{\arraystretch}{1.14}
\caption{Principal fields in \textsc{HalluTruthQA-4K}.}
\label{tab:data_fields}
\begin{tabularx}{\linewidth}{@{}
  >{\raggedright\arraybackslash}p{0.31\linewidth}
  >{\raggedright\arraybackslash}X@{}}
\toprule
\rowcolor{black!10}
\textbf{Field} & \textbf{Description} \\
\midrule

\rowcolor{paperlinkblue!7}
\multicolumn{2}{@{}l@{}}{\textbf{Core instance fields}} \\
\addlinespace[1pt]

\textbf{\texttt{id}}
& Unique instance identifier. \\

\rowcolor{gray!4}
\textbf{\texttt{question}}
& Arabic factual question. \\

\textbf{\texttt{generated\_answer}}
& Free-form response associated with the instance. \\

\rowcolor{gray!4}
\textbf{\texttt{gold\_answer}}
& Expert-verified reference answer. \\

\textbf{\texttt{generator\_model}}
& Identifier of the response-generation model. \\

\rowcolor{gray!4}
\textbf{\texttt{label}}
& \texttt{hallucination} or \texttt{no\_hallucination}. \\

\addlinespace[2pt]
\rowcolor{paperlinkblue!7}
\multicolumn{2}{@{}l@{}}{\textbf{Localized hallucination annotations}} \\
\addlinespace[1pt]

\textbf{\texttt{hallucinations}}
& List of localized-error objects; present only for hallucinated responses. \\

\rowcolor{gray!4}
\quad\textbf{\texttt{span\_start}},
\textbf{\texttt{span\_end}}
& Start and exclusive end character offsets in
\texttt{generated\_answer}. \\

\quad\textbf{\texttt{hallucinated\_span}}
& Exact erroneous text segment. \\

\rowcolor{gray!4}
\quad\textbf{\texttt{explanation}}
& Human-written explanation of the error, including the factual correction
when applicable. \\

\addlinespace[2pt]
\rowcolor{paperlinkblue!7}
\multicolumn{2}{@{}l@{}}{\textbf{Factual-verification fields}} \\
\addlinespace[1pt]

\textbf{\texttt{options}}
& Six candidate answers containing one verified answer and five plausible
distractors. \\

\rowcolor{gray!4}
\textbf{\texttt{answer}}
& Identifier of the verified candidate option. \\

\bottomrule
\end{tabularx}
\end{table}

Table~\ref{tab:data_example} shows an example from the Islamic knowledge
domain. In the generated answer, the substring from character 23 up to, but
not including, character 25 is \textarabic{29}, illustrating the zero-based,
end-exclusive span convention.

\begin{table}[t]
\centering
\small
\setlength{\tabcolsep}{5pt}
\renewcommand{\arraystretch}{1.10}
\caption{Example \textsc{HalluTruthQA-4K} instance from the Islamic knowledge
domain.}
\label{tab:data_example}
\begin{tabularx}{\linewidth}{@{}
  >{\raggedright\arraybackslash}p{0.29\linewidth}
  >{\raggedright\arraybackslash}X@{}}
\toprule
\textbf{Field} & \textbf{Value} \\
\midrule

\texttt{id} &
\texttt{islamicHallucination\_9} \\

\texttt{question} &
\textarabic{ما هي السورة القرآنية التي فيها موضعان للسجود؟} \\

\texttt{generated\_answer} &
\textarabic{سورة الحج، الآيتان 18 و29.} \\

\texttt{gold\_answer} &
\textarabic{سورة الحج، وموضعا السجود فيها في الآيتين 18 و77} \\

\texttt{generator\_model} &
\texttt{QCRI/Fanar-1-9B-Instruct} \\

\texttt{label} &
\texttt{hallucination} \\

\texttt{hallucinations} &
One localized error. \\

\quad\texttt{span\_start},
\texttt{span\_end} &
\texttt{23}, \texttt{25} \\

\quad\texttt{hallucinated\_span} &
\textarabic{29} \\

\quad\texttt{explanation} &
\textarabic{الإجابة الأساسية صحيحة؛ السورة التي فيها موضعان للسجود هي سورة الحج، لكن موضع السجدة الثانية ليس الآية 29، بل الآية 77.} \\

\texttt{options} &
A: \textarabic{السورة القرآنية التي فيها موضعان للسجود هي سورة الحج، وموضعا السجود فيها في الآيتين 18 و77.}\newline
B: \textarabic{السورة القرآنية التي فيها موضعان للسجود هي سورة الحج، وموضعا السجود فيها في الآيتين 18 و29.}\newline
C: \textarabic{السورة القرآنية التي فيها موضعان للسجود هي سورة السجدة، وموضعا السجود فيها في الآيتين 15 و16.}\newline
D: \textarabic{السورة القرآنية التي فيها موضعان للسجود هي سورة فصلت، وموضعا السجود فيها في الآيتين 37 و38.}\newline
E: \textarabic{السورة القرآنية التي فيها موضعان للسجود هي سورة النجم، وموضعا السجود فيها في الآيتين 62 و63.}\newline
F: \textarabic{السورة القرآنية التي فيها موضعان للسجود هي سورة العلق، وموضعا السجود فيها في الآيتين 19 و20.} \\

\texttt{answer} &
\texttt{A} \\

\bottomrule
\end{tabularx}
\end{table}

\FloatBarrier

\subsection{Hallucination Taxonomy}
\label{sec:hallucination_taxonomy}

We analyze the hallucination errors observed in the responses generated by
\textsc{Fanar-1-9B-Instruct} and organize them into a two-level taxonomy
comprising five macro-types and 22 micro-types. The  macro-types are \textit{Factual Contradiction},
\textit{Context Inconsistency}, \textit{Logical Inconsistency},
\textit{Factual Fabrication}, and
\textit{Nonsensical or Irrelevant Response}. Together, they distinguish
incorrect factual values, faulty grounding, invalid reasoning, invented
factual content, and responses that fail to provide a usable factual answer. 
Because a response may contain several independent errors, taxonomy labels
are assigned and analyzed at the span level rather than at the response
level. This allows a single response to contain multiple error types, such
as an incorrect entity combined with a fabricated source. Table~\ref{tab:macro_micro_definitions} defines the macro-types and their
corresponding micro-types. Macro-types capture the general nature of the
failure, while micro-types provide a more precise description of the
erroneous span.

\begin{figure*}[t]
    \centering
    \includegraphics[width=\textwidth]{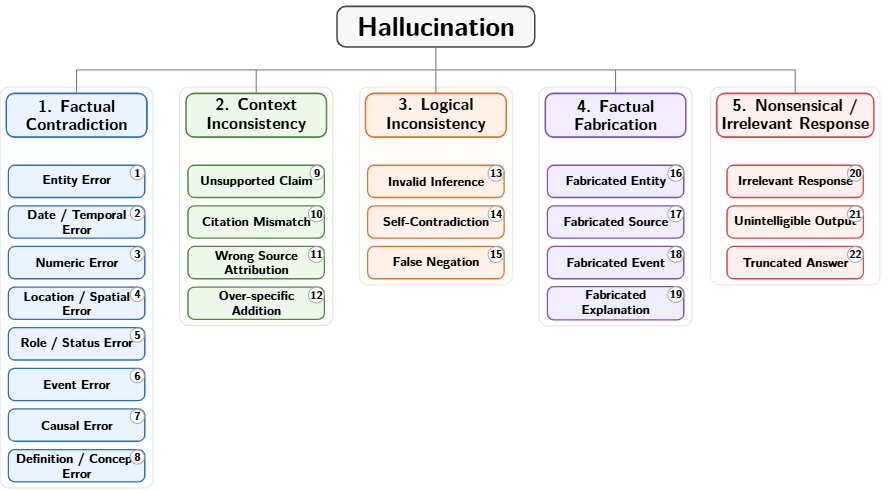}
    \caption{Two-level hallucination taxonomy. Macro-types represent broad failure categories,
    while micro-types identify the specific error affecting an annotated
    span.}
    \label{fig:hallucination_taxonomy}
\end{figure*}

\providecolor{fcLine}{HTML}{1E6ACB}
\providecolor{fcFill}{HTML}{EAF3FF}
\providecolor{ciLine}{HTML}{2E7D32}
\providecolor{ciFill}{HTML}{ECF8EC}
\providecolor{liLine}{HTML}{E86E1C}
\providecolor{liFill}{HTML}{FFF0E6}
\providecolor{ffLine}{HTML}{6F42C1}
\providecolor{ffFill}{HTML}{F3ECFF}
\providecolor{nrLine}{HTML}{D92D20}
\providecolor{nrFill}{HTML}{FFF1F0}

Across the annotated spans, \textit{Factual Contradiction} is the most
frequent macro-type (58.9\%), followed by
\textit{Context Inconsistency} (27.4\%) and
\textit{Factual Fabrication} (11.1\%).
\textit{Nonsensical or Irrelevant Response} and
\textit{Logical Inconsistency} are less frequent, accounting for
1.5\% and 1.1\%, respectively. The distribution of hallucination types also varies across domains.
\textit{Factual Contradiction} accounts for 79.8\% of the hallucinated
spans in history, 92.6\% in science, and 59.1\% in geography.
Islamic knowledge shows a different pattern, where
\textit{Context Inconsistency} accounts for 60.9\% of the hallucinated
spans. Geography shows the highest proportion of
\textit{Factual Fabrication} (28.7\%).   Table~\ref{tab:taxonomy_examples} provides  examples of
hallucination  observed in the corpus.

\begin{table}[p]
\centering
\scriptsize
\setlength{\tabcolsep}{3.8pt}
\renewcommand{\arraystretch}{1.17}
\caption{Macro-types, micro-types, and definitions in the hallucination
taxonomy.}
\label{tab:macro_micro_definitions}

\begin{tabularx}{\linewidth}{@{}
>{\raggedright\arraybackslash}p{0.20\linewidth}
>{\raggedright\arraybackslash}p{0.17\linewidth}
>{\raggedright\arraybackslash}X@{}}

\toprule
\textbf{Macro-type} & \textbf{Micro-type} & \textbf{Definition} \\
\midrule

\rowcolor{fcFill}
\textbf{\textcolor{fcLine}{Factual Contradiction}}\newline
\textit{\textcolor{fcLine}{Wrong factual value}}
& Entity Error
& The response gives an incorrect person, country, organization, object, or
named entity instead of the verified answer. \\
\cmidrule(l){2-3}

\rowcolor{fcFill}
& Date / Temporal Error
& The response gives an incorrect date, year, period, duration,
chronological order, or temporal relation. \\
\cmidrule(l){2-3}

\rowcolor{fcFill}
& Numeric Error
& The response gives an incorrect number, quantity, percentage, distance,
length, count, or measurement. \\
\cmidrule(l){2-3}

\rowcolor{fcFill}
& Location / Spatial Error
& The response gives an incorrect place, region, country, spatial relation,
or geographical location. \\
\cmidrule(l){2-3}

\rowcolor{fcFill}
& Role / Status Error
& The response assigns an incorrect role, title, status, function,
affiliation, or relation to an entity. \\
\cmidrule(l){2-3}

\rowcolor{fcFill}
& Event Error
& The response describes a real or expected event incorrectly, assigns it
to the wrong actor, time, place, or outcome, or confuses it with another
event. \\
\cmidrule(l){2-3}

\rowcolor{fcFill}
& Causal Error
& The response gives an incorrect cause, effect, motivation, or explanatory
relation between facts. \\
\cmidrule(l){2-3}

\rowcolor{fcFill}
& Definition / Concept Error
& The response confuses related concepts or gives a definition that
contradicts the verified answer. \\

\midrule

\rowcolor{ciFill}
\textbf{\textcolor{ciLine}{Context Inconsistency}}\newline
\textit{\textcolor{ciLine}{Wrong evidence or grounding}}
& Unsupported Claim
& The response adds a claim that remains related to the question but is not
supported by the verified evidence. \\
\cmidrule(l){2-3}

\rowcolor{ciFill}
& Citation Mismatch
& The response links a claim to the wrong citation, verse, hadith, book,
source, or evidential reference. \\
\cmidrule(l){2-3}

\rowcolor{ciFill}
& Wrong Source Attribution
& The response attributes a statement to a source, scholar, text, or
authority that does not support that statement. \\
\cmidrule(l){2-3}

\rowcolor{ciFill}
& Over-specific Addition
& The response adds details that are more specific than what the verified
evidence allows, without necessarily inventing a new source or event. \\

\midrule

\rowcolor{liFill}
\textbf{\textcolor{liLine}{Logical Inconsistency}}\newline
\textit{\textcolor{liLine}{Wrong reasoning}}
& Invalid Inference
& The response uses related information but draws a conclusion that does not
logically follow from it. \\
\cmidrule(l){2-3}

\rowcolor{liFill}
& Self-Contradiction
& The response contains two or more claims that are mutually incompatible
within the same answer. \\
\cmidrule(l){2-3}

\rowcolor{liFill}
& False Negation
& The response incorrectly denies a true relation or states the opposite of
the verified answer. \\

\midrule

\rowcolor{ffFill}
\textbf{\textcolor{ffLine}{Factual Fabrication}}\newline
\textit{\textcolor{ffLine}{Invented factual content}}
& Fabricated Entity
& The response introduces a non-attested entity and presents it as factual. \\
\cmidrule(l){2-3}

\rowcolor{ffFill}
& Fabricated Source
& The response invents a source, citation, reference, verse, hadith, book,
or authority that does not exist or is not verified. \\
\cmidrule(l){2-3}

\rowcolor{ffFill}
& Fabricated Event
& The response describes an event that is not supported by the verified
sources and appears to be invented. \\
\cmidrule(l){2-3}

\rowcolor{ffFill}
& Fabricated Explanation
& The response gives an explanation that sounds specific or authoritative
but is not grounded in the verified evidence. \\

\midrule

\rowcolor{nrFill}
\textbf{\textcolor{nrLine}{Nonsensical / Irrelevant Response}}\newline
\textit{\textcolor{nrLine}{No usable factual answer}}
& Irrelevant Response
& The response is off-topic or does not address the question in a meaningful
way. \\
\cmidrule(l){2-3}

\rowcolor{nrFill}
& Unintelligible Output
& The response is too unclear, incoherent, or broken to be interpreted as a
valid factual answer. \\
\cmidrule(l){2-3}

\rowcolor{nrFill}
& Truncated Answer
& The response is incomplete or stops before providing a verifiable answer. \\

\bottomrule
\end{tabularx}
\end{table}

\begin{table}[!h]
\centering
\small
\setlength{\tabcolsep}{5pt}
\renewcommand{\arraystretch}{1.15}

\caption{Examples of hallucination types. }
\label{tab:taxonomy_examples}

\begin{tabularx}{\textwidth}{@{}
>{\raggedright\arraybackslash}p{2.9cm}
>{\raggedright\arraybackslash}X
>{\raggedright\arraybackslash}X
>{\raggedright\arraybackslash}X@{}}

\toprule
\textbf{Type} &
\textbf{Question} &
\textbf{Generated response / Gold answer} &
\textbf{Explanation} \\
\midrule

\textbf{Factual Contradiction}\newline
Entity Error
&
\textarabic{من هو الصحابي الذي دعا له النبي صلى الله عليه وسلم بقوله:
اللهم فقهه في الدين؟}
&
\textbf{Pred.:} \textarabic{\textbf{معاذ بن جبل} رضي الله عنه}\newline
\textbf{Gold:} \textarabic{عبد الله بن عباس}
&
The response identifies the wrong Companion. The verified answer is
Abdullah Ibn Abbas. \\

\addlinespace[3pt]

\textbf{Factual Contradiction}\newline
Numeric Error
&
\textarabic{ما هي السورة القرآنية التي فيها موضعان للسجود؟}
&
\textbf{Pred.:} \textarabic{سورة الحج، الآيتان 18 و\textbf{29}.}\newline
\textbf{Gold:} \textarabic{سورة الحج، الآيتان 18 و77.}
&
The surah is correctly identified, but the second verse number is
incorrect: it is verse 77 rather than 29. \\

\addlinespace[3pt]

\textbf{Context Inconsistency}\newline
Wrong Source Attribution
&
\textarabic{أشار القرآن الكريم إلى عدة مهن كانت تزاول في الأمم السابقة،
منها مهنة بناء السفن، فما الآيات التي أشارت إلى هذه المهنة؟
ومن هو النبي الذي زاول هذه المهنة؟}
&
\textbf{Pred.:}
\textarabic{الآية 14 من سورة هود، موسى عليه السلام كان يزاول مهنة بناء
السفن، الآية 41 من سورة الأنبياء، الآية 40 من سورة هود، النبي الذي
زاول مهنة بناء السفن هو موسى عليه السلام.}
\newline
\textbf{Gold:}
\textarabic{هود: 37، 38، نوح عليه السلام}
&
The response gives incorrect Qur'anic references and attributes
shipbuilding to the wrong prophet. \\

\addlinespace[3pt]

\textbf{Factual Contradiction}\newline
Date / Temporal Error
&
\textarabic{ما المدينة التي تسمى ``عاصمة الثقافة العربية 2024''؟}
&
\textbf{Pred.:}
\textarabic{مدينة سطيف الجزائرية}
\newline
\textbf{Gold:}
\textarabic{المدينة التي حملت لقب ``عاصمة الثقافة العربية لعام 2024''
هي مدينة طرابلس اللبنانية. كانت مقررة أصلاً لعام 2023، لكنها تأجلت
بسبب الأزمة الاقتصادية اللبنانية. أما مدينة سطيف الجزائرية فقد حملت
لقب ``عاصمة الثقافة العربية'' في عام 2007م.}
&
The response incorrectly selects Setif, Algeria. The verified city for
2024 is Tripoli, Lebanon, while Setif received the title in 2007. \\

\bottomrule
\end{tabularx}
\end{table}
\section{Corpus Statistics and Analysis}
\label{sec:analysis}

We examine the main characteristics of \textsc{HalluTruthQA-4K}, including
how hallucinations are distributed across domains, the number and length of
localized errors, and how hallucination frequency varies with question and
response length.

\paragraph{Label and span distribution.}
\label{sec:distribution_analysis}

The corpus contains 1{,}789 hallucinated responses (44.7\%) and 2{,}211
non-hallucinated responses (55.3\%). Hallucination rates vary across
domains. Islamic knowledge has the highest rate at 49.7\%,
followed by geography at 47.5\%, science at 46.4\%, and history at 35.3\%.
A chi-squared test shows a significant association between domain and
hallucination label ($\chi^{2}(3)=50.19$, $p<10^{-10}$). The corresponding
Cram'er's $V$ is 0.112, indicating a small but systematic association
between domain and hallucination frequency.

\begin{figure}[!h]
\centering
\includegraphics[width=0.90\linewidth]
{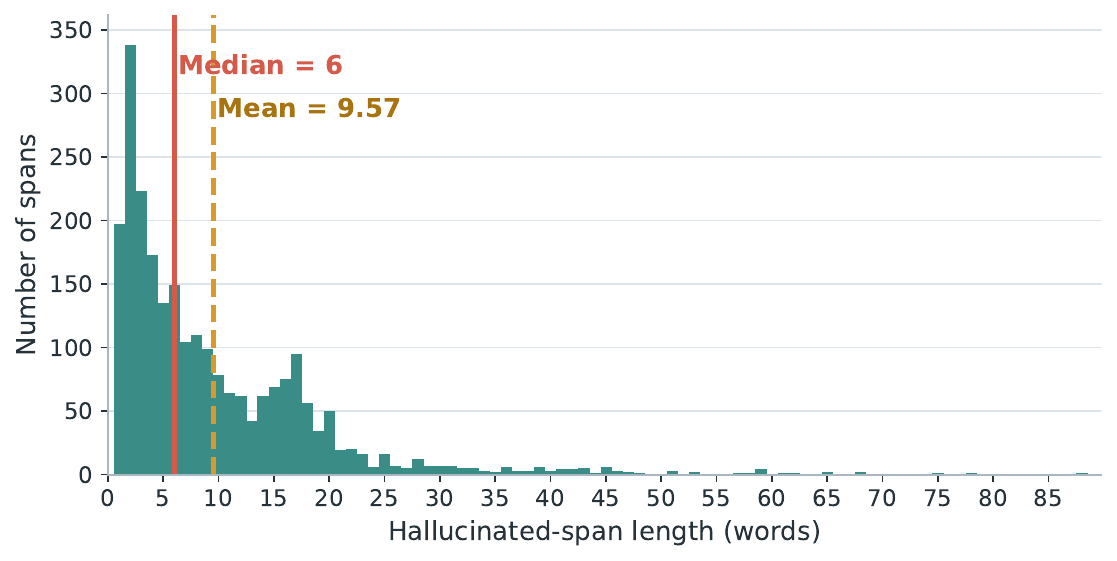}
\caption{Distribution of hallucinated-span lengths across the corpus.}
\label{fig:span_length}
\end{figure}

Across the 1{,}789 hallucinated responses, the corpus contains 2{,}419
localized erroneous spans. Every hallucinated response contains at least
one annotated span, while 396 responses (22.1\%) contain more than one
localized error.
The median hallucinated span contains 6 whitespace-separated words, while
the mean is 9.57 words. A total of 197 spans (8.1\%) contain only one word.
The distribution is concentrated on relatively short spans but has a long
tail, with some errors extending over substantially longer segments. This
pattern motivates character-level localization, since many responses
contain a localized factual error within an otherwise plausible answer
(Figure~\ref{fig:span_length}).

\paragraph{Question and response length.}
\label{sec:length_analysis}

\begin{figure}[!h]
\centering
\includegraphics[width=0.80\linewidth]
{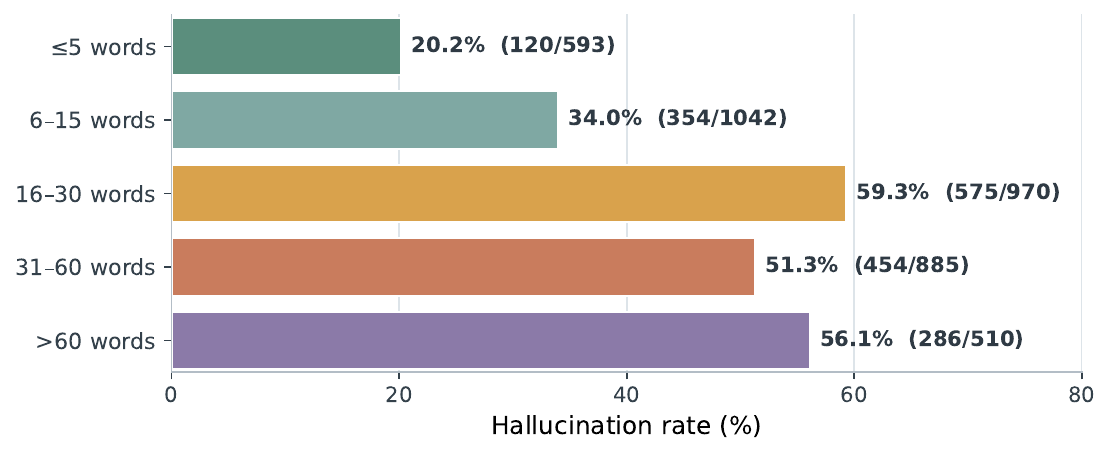}
\caption{Hallucination rate by generated-response length.}
\label{fig:len_vs_rate}
\end{figure}

\begin{figure}[!h]
\centering
\includegraphics[width=0.80\linewidth]
{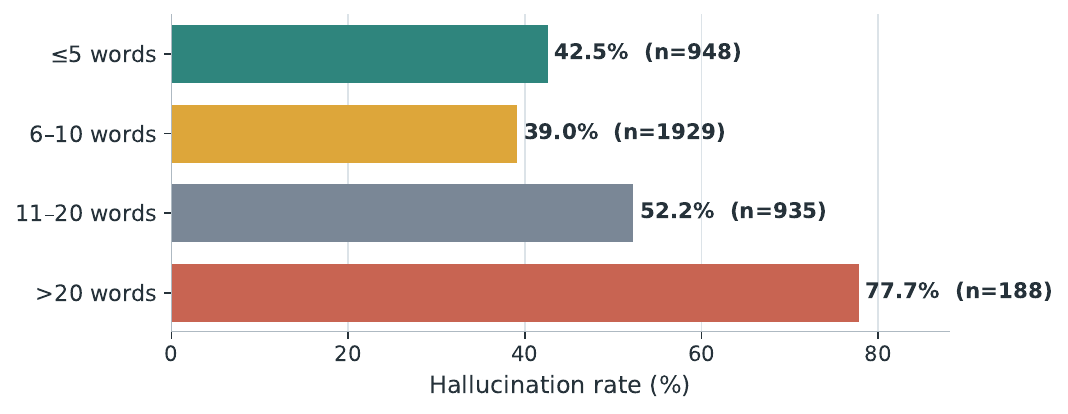}
\caption{Hallucination rate by question length.}
\label{fig:qlen_vs_rate}
\end{figure}

Hallucinated responses are longer on average than non-hallucinated
responses, with mean lengths of 36.15 and 26.79 whitespace-separated words,
respectively. Figure~\ref{fig:len_vs_rate} shows a non-monotonic relationship
between response length and hallucination rate.
Responses containing at most five words have a hallucination rate of 20.2\%.
The rate increases to 34.0\% for responses containing 6--15 words and
59.3\% for those containing 16--30 words. It remains high for longer
responses, reaching 51.3\% for 31--60 words and 56.1\% for responses
longer than 60 words. These results show that very short responses are less
likely to contain hallucinated content, whereas longer responses generally
involve more factual statements and therefore more opportunities for error.
However, the relationship is not strictly monotonic, indicating that
response length alone is not a reliable indicator of factual correctness.

Questions associated with hallucinated responses are also longer on
average, with mean lengths of 10.52 words compared with 8.38 words for
non-hallucinated responses. The hallucination rate is 42.5\% (403/948) for
questions containing at most five words and 39.0\% (752/1{,}929) for
questions containing 6--10 words. It increases to 52.2\% (488/935) for
questions containing 11--20 words and to 77.7\% (146/188) for questions
longer than 20 words (Figure~\ref{fig:qlen_vs_rate}).

Longer questions may contain more entities, constraints, temporal
relations, or requested facts, increasing the amount of information that
must be handled in the response. This result should nevertheless be
interpreted as an association rather than a causal effect, particularly
because the longest question-length group contains fewer instances than
the other groups.


\FloatBarrier

\section{Conclusion}
\label{sec:conclusion}

We introduced \textsc{HalluTruthQA-4K}, an expert-annotated Arabic resource
for evaluating factual reliability in knowledge-intensive question answering.
The corpus contains 4{,}000 instances across Islamic knowledge, history,
science, and geography, and combines response-level hallucination labels,
exact character-level error spans, human-written explanations, hierarchical
error types, verified reference answers, and six-option factual verification.
This design supports complementary evaluation of hallucination detection,
error localization, explanation, and factual answer selection within the same
QA instances.

The dataset was constructed through a multi-stage process involving
expert-prepared questions and reference answers, controlled response
generation, manual candidate-answer construction, independent verification,
and final adjudication. The resulting annotations make it possible to study
not only whether a response contains a factual error, but also where the error
occurs, how it can be characterized, why it is incorrect, and what the
verified answer should be. \textsc{HalluTruthQA-4K} also serves as the official
dataset for Track~2 of the HalluScoring 2026 shared task, providing a common
evaluation setting for comparing systems.

The corpus has several limitations. All generated responses were produced by
a single Arabic-centric model, and the four selected domains do not cover the
full range of Arabic varieties, topics, and knowledge sources. In addition,
multiple-choice verification evaluates factual selection from predefined
candidates and does not fully capture open-ended correction. Future work will
extend the resource with responses from additional model families, broader
domains and Arabic varieties, and further experiments on cross-model and
cross-domain generalization. We release \textsc{HalluTruthQA-4K} to support
reproducible research on detecting, localizing, explaining, and verifying
factual errors in Arabic LLM outputs.
\section{Appendix}
\appendix

\section{Representative Source Materials}
\label{app:sources}

The following sources are representative of the materials consulted during
question preparation and reference verification. The list is illustrative
rather than exhaustive.
\\
\\
\textbf{Islamic knowledge.}
Representative sources included IslamWeb, IslamQA, and Arabic reference books
such as:

\begin{itemize}
    \setlength{\itemsep}{1pt}
    \setlength{\parskip}{0pt}
    \setlength{\parsep}{0pt}
    \setlength{\topsep}{0pt}
    \setlength{\partopsep}{0pt}
    \item \textarabic{1213 سؤال وجواب في القرآن الكريم}
    (\textit{1,213 Questions and Answers on the Holy Qur'an}),
    by Ahmad bin Ali Abu Islam;

    \item \textarabic{200 سؤال وجواب في العقيدة الإسلامية}
    (\textit{200 Questions and Answers on Islamic Creed}),
    by Hafiz bin Ahmad Al-Hakami;

    \item \textarabic{السيرة النبوية في سؤال وجواب}
    (\textit{The Prophetic Biography in Questions and Answers}),
    by Abu Abdullah Muhammad Ali Samak;

    \item \textarabic{أكثر من 1500 سؤال وجواب في القرآن الكريم}
    (\textit{More Than 1,500 Questions and Answers on the Holy Qur'an}),
    by Muhsin Hussein Al-Ghamdi;

    \item \textarabic{الموسوعة القرآنية}
    (\textit{The Qur'anic Encyclopedia}).
\end{itemize}

\textbf{History.}
Representative sources included
\textarabic{أطلس تاريخ الإسلام}
(\textit{Atlas of Islamic History}) and
\textarabic{المفصل في تاريخ العرب قبل الإسلام}
(\textit{The Detailed History of the Arabs Before Islam}).
\\
\\
\textbf{Geography.}
Representative sources included
\textarabic{أطلس الوطن العربي}
(\textit{Atlas of the Arab World}),
\textarabic{جغرافية العالم الإسلامي}
(\textit{Geography of the Islamic World}), and
\textarabic{الموسوعة العربية العالمية}
(\textit{The Global Arabic Encyclopedia}).
\\
\\
\textbf{Science.}
Representative sources included
\textarabic{موسوعة العلوم والتقانات}
(\textit{Encyclopedia of Science and Technology}) and
\textarabic{الموسوعة العلمية الميسرة}
(\textit{The Concise Scientific Encyclopedia}).

\bibliographystyle{plainnat}
\bibliography{references}

\begin{thebibliography}{27}
\providecommand{\natexlab}[1]{#1}
\providecommand{\url}[1]{\texttt{#1}}
\expandafter\ifx\csname urlstyle\endcsname\relax
  \providecommand{\doi}[1]{doi: #1}\else
  \providecommand{\doi}{doi: \begingroup \urlstyle{rm}\Url}\fi

\bibitem[Abdaljalil et~al.(2026)Abdaljalil, Sharma, Serpedin, and Kurban]{abdaljalil2026halluverse}
Samir Abdaljalil, Parichit Sharma, Erchin Serpedin, and Hasan Kurban.
\newblock {Halluverse-M$^3$}: A multitask multilingual benchmark for hallucination in {LLM}s.
\newblock \emph{arXiv preprint arXiv:2602.06920}, 2026.

\bibitem[Abdelaal et~al.(2026)Abdelaal, Haffar, Fawzi, and Magdy]{abdelaal2026islamicmmlu}
Ali Abdelaal, Mohammed Nader~Al Haffar, Mahmoud Fawzi, and Walid Magdy.
\newblock {IslamicMMLU}: A benchmark for evaluating {LLM}s on islamic knowledge.
\newblock \emph{arXiv preprint arXiv:2603.23750}, 2026.

\bibitem[Alansari and Luqman(2025)]{alansari2025arahallueval}
Aisha Alansari and Hamzah Luqman.
\newblock {A}ra{H}allu{E}val: A fine-grained hallucination evaluation framework for {A}rabic {LLM}s.
\newblock In \emph{Proceedings of The Second Arabic Natural Language Processing Conference}, pages 148--161, Suzhou, China, November 2025. Association for Computational Linguistics.
\newblock \doi{10.18653/v1/2025.arabicnlp-main.12}.
\newblock URL \url{https://aclanthology.org/2025.arabicnlp-main.12/}.

\bibitem[Alansari and Luqman(2026{\natexlab{a}})]{alansari2026halluscore}
Aisha Alansari and Hamzah Luqman.
\newblock {HalluScore}: Large language model hallucination question answering benchmark.
\newblock \emph{arXiv preprint arXiv:2605.17007}, 2026{\natexlab{a}}.

\bibitem[Alansari and Luqman(2026{\natexlab{b}})]{alansari2026survey}
Aisha Alansari and Hamzah Luqman.
\newblock Large language models hallucination: A comprehensive survey.
\newblock \emph{arXiv preprint arXiv:2510.06265}, 2026{\natexlab{b}}.
\newblock URL \url{https://arxiv.org/abs/2510.06265}.

\bibitem[Alwajih et~al.(2025)Alwajih, El~Mekki, Mubarak, Hawasly, Mohamed, and Abdul-Mageed]{alwajih2025palmx}
Fakhraddin Alwajih, Abdellah El~Mekki, Hamdy Mubarak, Majd Hawasly, Abubakr Mohamed, and Muhammad Abdul-Mageed.
\newblock {PalmX} 2025: The first shared task on benchmarking {LLM}s on arabic and islamic culture.
\newblock In \emph{Proceedings of The Third Arabic Natural Language Processing Conference: Shared Tasks}, pages 774--789, 2025.

\bibitem[Bhatia et~al.(2026)Bhatia, Mubarak, Jarrar, Mikros, Zaraket, Alhirthani, Al-Khatib, Cochrane, Darwish, Yahiaoui, and Alam]{bhatia2026islamicfaithqa}
Gagan Bhatia, Hamdy Mubarak, Mustafa Jarrar, George Mikros, Fadi Zaraket, Mahmoud Alhirthani, Mutaz Al-Khatib, Logan Cochrane, Kareem Darwish, Rashid Yahiaoui, and Firoj Alam.
\newblock From {RAG} to agentic {RAG} for faithful islamic question answering.
\newblock \emph{arXiv preprint arXiv:2601.07528}, 2026.

\bibitem[Bouchekif et~al.(2025)Bouchekif, Rashwani, Mohamed, Alkhatib, Sbahi, Gaben, Zaghouani, Erbad, and Ghaly]{qias2025}
Abdessalam Bouchekif, Samer Rashwani, Emad Soliman~Ali Mohamed, Mutaz Alkhatib, Heba Sbahi, Shahd Gaben, Wajdi Zaghouani, Aiman Erbad, and Mohammed Ghaly.
\newblock {QIAS} 2025: Overview of the shared task on islamic inheritance reasoning and knowledge assessment.
\newblock In \emph{Proceedings of The Third Arabic Natural Language Processing Conference: Shared Tasks}, pages 851--860, Suzhou, China, November 2025. Association for Computational Linguistics.
\newblock ISBN 979-8-89176-356-2.
\newblock \doi{10.18653/v1/2025.arabicnlp-sharedtasks.117}.
\newblock URL \url{https://aclanthology.org/2025.arabicnlp-sharedtasks.117/}.

\bibitem[Bouchekif et~al.(2026{\natexlab{a}})Bouchekif, Eltanbouly, Rashwani, Gaben, Al-Khatib, Sbahi, Mohamed, and Ghaly]{qias2026}
Abdessalam Bouchekif, Somaya Eltanbouly, Samer Rashwani, Shahd Gaben, Mutaz Al-Khatib, Heba Sbahi, Emad Mohamed, and Mohammed Ghaly.
\newblock {QIAS} 2026: Overview of the shared task on islamic inheritance reasoning.
\newblock \emph{arXiv preprint arXiv:2606.13756}, 2026{\natexlab{a}}.

\bibitem[Bouchekif et~al.(2026{\natexlab{b}})Bouchekif, Gaben, Rashwani, Eltanbouly, Al-Khatib, Sbahi, Ghaly, and Mohamed]{bouchekif2026mawarith}
Abdessalam Bouchekif, Shahd Gaben, Samer Rashwani, Somaya Eltanbouly, Mutaz Al-Khatib, Heba Sbahi, Mohammed Ghaly, and Emad Mohamed.
\newblock {MAWARITH}: A dataset and benchmark for legal inheritance reasoning with {LLM}s.
\newblock \emph{arXiv preprint arXiv:2603.07539}, 2026{\natexlab{b}}.

\bibitem[Bouchekif et~al.(2026{\natexlab{c}})Bouchekif, Zighem, Bekhouche, Telli, Eltanbouly, Gaben, Sbahi, Rashwani, Al-Khatib, Mohamed, et~al.]{bouchekif2026hallutruthqa}
Abdessalam Bouchekif, Mohammed-En-Nadhir Zighem, Salah~Eddine Bekhouche, Hichem Telli, Somaya Eltanbouly, Shahd Gaben, Heba Sbahi, Samer Rashwani, Mutaz Al-Khatib, Emad Mohamed, et~al.
\newblock {HalluTruthQA}: A fine-grained benchmark for hallucination detection, localization, and explanation in arabic question answering.
\newblock \emph{arXiv preprint arXiv:2607.20219}, 2026{\natexlab{c}}.

\bibitem[Chen et~al.(2023)Chen, Zhao, Zhang, Chern, Gao, Liu, and He]{chen2023felm}
Shiqi Chen, Yiran Zhao, Jinghan Zhang, I-Chun Chern, Siyang Gao, Pengfei Liu, and Junxian He.
\newblock {FELM}: Benchmarking factuality evaluation of large language models.
\newblock In \emph{Thirty-seventh Conference on Neural Information Processing Systems Datasets and Benchmarks Track}, 2023.
\newblock URL \url{https://arxiv.org/abs/2310.00741}.

\bibitem[Elchafei and Abu-Elkheir(2025)]{elchafei2025hallucination}
Passant Elchafei and Mervat Abu-Elkheir.
\newblock Hallucination detectives at {S}em{E}val-2025 task 3: Span-level hallucination detection for {LLM}-generated answers.
\newblock In \emph{Proceedings of the 19th International Workshop on Semantic Evaluation ({SemEval}-2025)}, pages 601--606, Vienna, Austria, July 2025. Association for Computational Linguistics.
\newblock URL \url{https://aclanthology.org/2025.semeval-1.84/}.

\bibitem[{Fanar Team} et~al.(2025){Fanar Team}, Abbas, Ahmad, Alam, Altinisik, Asgari, Boshmaf, Boughorbel, Chawla, Chowdhury, Dalvi, Darwish, Durrani, Elfeky, Elmagarmid, Eltabakh, Fatehkia, Fragkopoulos, Hasanain, Hawasly, Husaini, Jung, Lucas, Magdy, Messaoud, Mohamed, Mohiuddin, Mousi, Mubarak, Musleh, Naeem, Ouzzani, Popovic, Sadeghi, Sencar, Shinoy, Sinan, Zhang, Ali, El~Kheir, Ma, and Ruan]{fanarteam2025fanar}
{Fanar Team}, Ummar Abbas, Mohammad~Shahmeer Ahmad, Firoj Alam, Enes Altinisik, Ehsannedin Asgari, Yazan Boshmaf, Sabri Boughorbel, Sanjay Chawla, Shammur Chowdhury, Fahim Dalvi, Kareem Darwish, Nadir Durrani, Mohamed Elfeky, Ahmed Elmagarmid, Mohamed Eltabakh, Masoomali Fatehkia, Anastasios Fragkopoulos, Maram Hasanain, Majd Hawasly, Mus'ab Husaini, Soon-Gyo Jung, Ji~Kim Lucas, Walid Magdy, Safa Messaoud, Abubakr Mohamed, Tasnim Mohiuddin, Basel Mousi, Hamdy Mubarak, Ahmad Musleh, Zan Naeem, Mourad Ouzzani, Dorde Popovic, Amin Sadeghi, Husrev~Taha Sencar, Mohammed Shinoy, Omar Sinan, Yifan Zhang, Ahmed Ali, Yassine El~Kheir, Xiaosong Ma, and Chaoyi Ruan.
\newblock {Fanar}: An arabic-centric multimodal generative {AI} platform, 2025.
\newblock URL \url{https://arxiv.org/abs/2501.13944}.

\bibitem[Gaben et~al.(2026)Gaben, Sbahi, Rashwani, Bouchekif, Al-Khatib, Mohamed, Eltanbouly, and Ghaly]{gaben2026islamicturathbench}
Shahd Gaben, Heba Sbahi, Samer Rashwani, Abdessalam Bouchekif, Mutaz Al-Khatib, Emad Mohamed, Somaya Eltanbouly, and Mohammed Ghaly.
\newblock Islamicturathbench: A multi-task, multi-discipline benchmark for evaluating large language models on the islamic scholarly tradition (turath).
\newblock \emph{arXiv preprint arXiv:2608.04703}, 2026.

\bibitem[Huang et~al.(2025{\natexlab{a}})Huang, Yu, Ma, Zhong, Feng, Wang, Chen, Peng, Feng, Qin, and Liu]{huang-etal-2025-survey}
Lei Huang, Weijiang Yu, Weitao Ma, Weihong Zhong, Zhangyin Feng, Haotian Wang, Qianglong Chen, Weihua Peng, Xiaocheng Feng, Bing Qin, and Ting Liu.
\newblock A survey on hallucination in large language models: Principles, taxonomy, challenges, and open questions.
\newblock \emph{ACM Transactions on Information Systems}, 43\penalty0 (2), 2025{\natexlab{a}}.
\newblock \doi{10.1145/3703155}.

\bibitem[Huang et~al.(2025{\natexlab{b}})Huang, Yu, Ma, Zhong, Feng, Wang, Chen, Peng, Feng, Qin, et~al.]{huang2025survey}
Lei Huang, Weijiang Yu, Weitao Ma, Weihong Zhong, Zhangyin Feng, Haotian Wang, Qianglong Chen, Weihua Peng, Xiaocheng Feng, Bing Qin, et~al.
\newblock A survey on hallucination in large language models: Principles, taxonomy, challenges, and open questions.
\newblock \emph{ACM Transactions on Information Systems}, 43\penalty0 (2):\penalty0 1--55, 2025{\natexlab{b}}.

\bibitem[Ji et~al.(2023)Ji, Lee, Frieske, Yu, Su, Xu, Ishii, Bang, Madotto, and Fung]{ji2023survey}
Ziwei Ji, Nayeon Lee, Rita Frieske, Tiezheng Yu, Dan Su, Yan Xu, Etsuko Ishii, Ye~Jin Bang, Andrea Madotto, and Pascale Fung.
\newblock Survey of hallucination in natural language generation.
\newblock \emph{ACM Computing Surveys}, 55\penalty0 (12):\penalty0 1--38, 2023.
\newblock \doi{10.1145/3571730}.

\bibitem[Li et~al.(2023)Li, Cheng, Zhao, Nie, and Wen]{li2023halueval}
Junyi Li, Xiaoxue Cheng, Xin Zhao, Jian-Yun Nie, and Ji-Rong Wen.
\newblock {H}alu{E}val: A large-scale hallucination evaluation benchmark for large language models.
\newblock In \emph{Proceedings of the 2023 Conference on Empirical Methods in Natural Language Processing}, pages 6449--6464, Singapore, December 2023. Association for Computational Linguistics.
\newblock \doi{10.18653/v1/2023.emnlp-main.397}.
\newblock URL \url{https://aclanthology.org/2023.emnlp-main.397/}.

\bibitem[Lin et~al.(2022)Lin, Hilton, and Evans]{lin2022truthfulqa}
Stephanie Lin, Jacob Hilton, and Owain Evans.
\newblock {T}ruthful{QA}: Measuring how models mimic human falsehoods.
\newblock In \emph{Proceedings of the 60th Annual Meeting of the Association for Computational Linguistics (Volume 1: Long Papers)}, pages 3214--3252, Dublin, Ireland, May 2022. Association for Computational Linguistics.
\newblock \doi{10.18653/v1/2022.acl-long.229}.
\newblock URL \url{https://aclanthology.org/2022.acl-long.229/}.

\bibitem[Min et~al.(2023)Min, Krishna, Lyu, Lewis, Yih, Koh, Iyyer, Zettlemoyer, and Hajishirzi]{min2023factscore}
Sewon Min, Kalpesh Krishna, Xinxi Lyu, Mike Lewis, Wen-tau Yih, Pang Koh, Mohit Iyyer, Luke Zettlemoyer, and Hannaneh Hajishirzi.
\newblock {FA}ct{S}core: Fine-grained atomic evaluation of factual precision in long form text generation.
\newblock In \emph{Proceedings of the 2023 Conference on Empirical Methods in Natural Language Processing}, pages 12076--12100, Singapore, December 2023. Association for Computational Linguistics.
\newblock \doi{10.18653/v1/2023.emnlp-main.741}.
\newblock URL \url{https://aclanthology.org/2023.emnlp-main.741/}.

\bibitem[Mishra et~al.(2024)Mishra, Asai, Balachandran, Wang, Neubig, Tsvetkov, and Hajishirzi]{mishra2024fava}
Abhika Mishra, Akari Asai, Vidhisha Balachandran, Yizhong Wang, Graham Neubig, Yulia Tsvetkov, and Hannaneh Hajishirzi.
\newblock Fine-grained hallucination detection and editing for language models.
\newblock \emph{arXiv preprint arXiv:2401.06855}, 2024.

\bibitem[Mohammed et~al.(2025)Mohammed, Ali, Ali, Majeed, and Mohamed]{mohammed2025aftina}
Marryam~Yahya Mohammed, Sama~Ayman Ali, Salma~Khaled Ali, Ayad~Abdul Majeed, and Ensaf~Hussein Mohamed.
\newblock Aftina: Enhancing stability and preventing hallucination in {AI}-based islamic fatwa generation using {LLM}s and {RAG}.
\newblock \emph{Neural Computing and Applications}, 37\penalty0 (25):\penalty0 20957--20982, 2025.

\bibitem[Mubarak et~al.(2024)Mubarak, Al-Khalifa, and Alkhalefah]{mubarak2024halwasa}
Hamdy Mubarak, Hend Al-Khalifa, and Khaloud~Suliman Alkhalefah.
\newblock Halwasa: Quantify and analyze hallucinations in large language models: {A}rabic as a case study.
\newblock In \emph{Proceedings of the 2024 Joint International Conference on Computational Linguistics, Language Resources and Evaluation ({LREC-COLING} 2024)}, pages 8008--8015, Torino, Italia, May 2024. ELRA and ICCL.
\newblock URL \url{https://aclanthology.org/2024.lrec-main.705/}.

\bibitem[Mubarak et~al.(2025)Mubarak, Malhas, Mansour, Mohamed, Fawzi, Hawasly, Elsayed, Darwish, and Magdy]{mubarak2025islamiceval}
Hamdy Mubarak, Rana Malhas, Watheq Mansour, Abubakr Mohamed, Mahmoud Fawzi, Majd Hawasly, Tamer Elsayed, Kareem~Mohamed Darwish, and Walid Magdy.
\newblock {IslamicEval} 2025: The first shared task of capturing {LLM}s hallucination in islamic content.
\newblock In \emph{Proceedings of The Third Arabic Natural Language Processing Conference: Shared Tasks}, pages 480--493, 2025.

\bibitem[Niu et~al.(2024)Niu, Wu, Zhu, Xu, Shum, Zhong, Song, and Zhang]{niu2024ragtruth}
Cheng Niu, Yuanhao Wu, Juno Zhu, Siliang Xu, KaShun Shum, Randy Zhong, Juntong Song, and Tong Zhang.
\newblock {RAGT}ruth: A hallucination corpus for developing trustworthy retrieval-augmented language models.
\newblock In \emph{Proceedings of the 62nd Annual Meeting of the Association for Computational Linguistics (Volume 1: Long Papers)}, pages 10862--10878, Bangkok, Thailand, August 2024. Association for Computational Linguistics.
\newblock \doi{10.18653/v1/2024.acl-long.585}.
\newblock URL \url{https://aclanthology.org/2024.acl-long.585/}.

\bibitem[Vazquez et~al.(2025)Vazquez, Mickus, Zosa, Vahtola, Tiedemann, Sinha, Segonne, Sanchez-Vega, Raganato, Libovick{\'y}, Karlgren, Ji, Helcl, Guillou, De~Gibert, Bengoetxea, Apidianaki, and Apidianaki]{vazquez2025mushroom}
Raul Vazquez, Timothee Mickus, Elaine Zosa, Teemu Vahtola, J{\"o}rg Tiedemann, Aman Sinha, Vincent Segonne, Fernando Sanchez-Vega, Alessandro Raganato, Jind{\v{r}}ich Libovick{\'y}, Jussi Karlgren, Shaoxiong Ji, Jind{\v{r}}ich Helcl, Liane Guillou, Ona De~Gibert, Jaione Bengoetxea, Joseph Apidianaki, and Marianna Apidianaki.
\newblock {S}em{E}val-2025 task 3: Mu-{SHROOM}, the multilingual shared-task on hallucinations and related observable overgeneration mistakes.
\newblock In \emph{Proceedings of the 19th International Workshop on Semantic Evaluation ({SemEval}-2025)}, pages 2472--2497, Vienna, Austria, July 2025. Association for Computational Linguistics.
\newblock URL \url{https://aclanthology.org/2025.semeval-1.322/}.

\end{thebibliography}

\end{document}